# Reliable Breast Cancer Molecular Subtype Prediction based on uncertainty-aware Bayesian Deep Learning by Mammography

Mohaddeseh Chegini [a], Ali Mahloojifar [a,*]

[a] Department of Biomedical Engineering, Faculty of Electrical and Computer Engineering, Tarbiat Modares University, Tehran, Iran

*Corresponding author.

Email: mahlooji@modares.ac.ir (Ali Mahloojifar)

Email: m.chegini@modares.ac.ir (Mohaddeseh Chegini)

**Abstract**

Breast cancer is a heterogeneous disease with different molecular subtypes, clinical behavior, treatment responses as well as survival outcomes. The development of a reliable, accurate, available and inexpensive method to predict the molecular subtypes using medical images plays an important role in the diagnosis and prognosis of breast cancer. Recently, deep learning methods have shown good performance in the breast cancer classification tasks using various medical images. Despite all that success, classical deep learning cannot deliver the predictive uncertainty. The uncertainty represents the validity of the predictions. Therefore, the high predicted uncertainty might cause a negative effect in the accurate diagnosis of breast cancer molecular subtypes. To overcome this, uncertainty quantification methods are used to determine the predictive uncertainty. Accordingly, in this study, we proposed an uncertainty-aware Bayesian deep learning model using the full mammogram images. In addition, to increase the performance of the multi-class molecular subtype classification task, we proposed a novel hierarchical classification strategy, named the two-stage classification strategy. The separate AUC of the proposed model for each subtype was 0.71, 0.75 and 0.86 for HER2-enriched, luminal and triple-negative classes, respectively. The proposed model not only has a comparable performance to other studies in the field of breast cancer molecular subtypes prediction, even using full mammography images, but it is also more reliable, due to quantify the predictive uncertainty.

**Keywords:** Breast cancer, Computer-aided Diagnosis System, Bayesian Deep learning, Uncertainty Quantification, Molecular Subtype, Mammography.

1. Introduction

Breast cancer is one of the most commonly diagnosed cancers in women, and the second leading cause of cancer death among women overall [1]. It is a heterogeneous disease with different molecular subtypes, clinical behavior, treatment responses as well as survival outcomes[2]. Molecular heterogeneity in breast cancer is known in terms of varying tumor markers expression of the estrogen receptor (ER), progesterone receptor (PR), human epidermal growth factor receptor 2 (HER2) and Ki-67 index. Determining the breast cancer molecular subtypes is an important prognostic factor and can facilitate individualized treatment because the therapies typically target these receptors. Depending on the expression level of certain receptors, breast cancer can be divided into various subtypes, such as the luminal (Luminal A (ER and/or PR positive, HER2 negative, low Ki-67 index) and Luminal B (ER and/or PR positive, HER2 negative with high Ki-67 or HER2 positive with any Ki-67 status)), human epidermal growth factor receptor 2 or HER2-enriched (ER and PR negative with HER2 positive), and triple-negative (TN) subtype (ER, PR, HER2 negative)[3].

At present, immunohistochemistry methods are mainly used to detect receptor expression status, which rely on small tissue samples of malignancy obtained through biopsy or surgery[4]. However, due to the spatial heterogeneity of breast tumors, this part of tumor tissue does not represent the complete features of the tumor. Additionally, there is temporal heterogeneity of breast tumors, in which the tumor features i.e., receptor status and molecular subtypes, may have changed over time due to the treatment [5]. Also, it is an expensive test, and not readily available in many developing and under-developed countries [6]. Therefore, there is a need for the development of a reliable, accurate, available and inexpensive alternative method to predict the molecular subtypes of breast cancer and provide personalized cancer treatment plans.

Recently, medical image technology and processing methods have made progress in using the radiomics approach to predict the molecular subtypes in breast cancer. Radiomics refers to extracting a large number of quantitative features from medical images and transforming them into high-dimensional data that can provide additional information for diagnosis, prognosis, and other forms of hidden important biological and medical information, which they are difficult to identify quantitatively by the human eye[7]. Some recent studies showed that the radiomics features obtained from magnetic resonance imaging (MRI), Mammography (MG) and Ultrasound (US) imaging can be associated with the molecular subtypes of breast cancer [3, 5-18]. Magnetic resonance imaging (MRI) has higher sensitivity than sonography and mammography [19], but it is an expensive examination and not widely available, especially in less developed countries [12]. Mammography is less expensive and widely used for breast cancer screening and diagnosis [20]. Therefore, the extraction of radiomic features for accurate and reliable prediction of molecular subtypes by routinely performed mammography will be valuable and important.

Recently, several advances in deep learning [21] have played a widespread role in medical image processing. Specifically, Convolutional Neural Networks (CNN), as a type of deep learning model, achieved proper results for breast lesion detection as well as classification [22-24]. Despite the advantages, deep learning models cannot quantify the uncertainty of the predictions and they are too confident about the outputs. Therefore, there are some challenges with unreliability.

In other words, many forms of uncertainty may arise in deep learning models. There are two main types of uncertainty, namely, aleatoric and epistemic uncertainties. The aleatoric or data uncertainty is introduced from noisy data such as measurement noise and it is an inherent property of the data. Also, the epistemic or model uncertainty occurs due to inadequate knowledge about which values of model parameters will be good at predicting new data, as well as, the general structure of the model. These scenarios may present the result in which it is difficult to determine whether a model is providing accurate predictions or just making random guesses[25-27]. Therefore, uncertainty information in the predicted results of a disease diagnosis systems, which make decisions that affect human life, is much more important. So, leveraging uncertainty information is imperative to ensure the safe use of these novel tools[28].

One of the most important methods to quantify uncertainty values is the probabilistic Bayesian approach. Bayesian deep learning (BDL) uses Bayesian probability theory in deep learning models to extract knowledge and learning from data, as well as, quantify all forms of uncertainty in the models. With regard to the posterior distribution, obtained by Bayes' theorem, the predictive probability distribution quantifies the uncertainty on its prediction [29, 30]. Practically, the true posterior cannot be evaluated analytically, hence an approximation is needed

[26]. The Monte Carlo Dropout method as a Bayesian approximation is used as an uncertainty quantification method. This method is a new framework that uses dropout in deep neural networks as approximate Bayesian inference in deep Gaussian processes. The Monte Carlo Dropout method can mitigate the problem of quantifying uncertainty in deep learning without sacrificing either computational complexity or test accuracy [31].

In this study, we propose a reliable bayesian deep learning model to classify breast tumor molecular subtypes which uses the Monte Carlo Dropout method to quantify the uncertainty in the predictions.

In addition, with the aim of increasing the performance in the tumor molecular subtypes multi-class classification task, we proposed a novel hierarchical classification strategy, named the two-stage classification strategy.

In summary, the main contributions of this study are as follows:

- To the best of the authors' knowledge, this is the first work where the full mammogram images were used to classify breast tumor molecular subtypes.
- We used a two-stage classification strategy for the tumor molecular subtypes multi-class classification task to increase the accuracy.
- We quantified the uncertainty in the deep learning model using the Monte Carlo Dropout technique as a Bayesian approximation.
- We showed that the Monte Carlo method in the bayesian deep learning model, in addition to quantify the uncertainty, also slightly increases the performance.
- The proposed uncertainty-aware bayesian deep learning-based model has a comparable performance to other studies in the field of breast cancer molecular subtypes prediction, even using full mammography images.
- To prognosis of breast cancer diagnosis in the early stages, it is very important to accurately diagnose the most dangerous type of cancer, known as the triple negative subtype,[20]. The proposed model demonstrates a proper performance in detecting the triple negative subtype.

**2. Related works**

Researchers present some machine learning or deep learning-based models using imaging modalities of the breast which some of them provide promising performance in breast tumor molecular subtypes prediction.

Zhang et al.[9] proposed a deep convolutional neural network for molecular subtype classification of the triple-negative, HER2 (+) and HR (+) subtypes using ultrasound images. Jiang et al.[15] used ResNet50 to classify luminal A, luminal B, HER2+ and triple-negative subtypes on ultrasound images. Zhang et al.[16] have proposed the ensemble decision approach for the classification of four class molecular subtypes using ultrasound and clinical features. Li et al.[32] developed a deep convolutional neural network model to classify luminal vs. non-luminal, Luminal A vs. non-luminal A and Triple-negative vs. non-triple-negative molecular subtypes of breast cancer from ultrasound images together with clinical information. Xu et al.[33] proposed a logistic regression algorithm to select molecular-related features and support vector machine models to predict biomarkers for discriminating the molecular biomarker profiling (ER, PR, HER2, and Ki-67).

Similarly, different machine learning and deep learning models have been proposed for breast tumor molecular subtype classification on MR images. Meng et al.[5] proposed a U-Net model and Gradient Tree Boosting for segmentation and classification using extracted shape and texture of the tumors and the clinical features into four types of molecular subtypes. Zhang et al.[34] used a conventional convolutional neural network and a recurrent convolutional neural network to differentiate three breast cancer molecular subtypes based on hormonal receptor (HR) and HER2 receptor. Ha et al.[35] proposed a convolutional neural network and the support vector machine to distinguish the four molecular subtypes of breast cancer.

Mammography, as the gold standard for breast cancer screening, has also been used by researchers in the field of breast tumor molecular subtype classification. Ma et al.[12] used quantitative radiomic features, including morphologic, grayscale statistic, and texture features and employed the Naive Bayes machine learning scheme to classify the molecular subtypes. They also used the least absolute shrinkage and selection operator (LASSO) method to select the most predictive features for the classifiers. Ueda et al.[36] proposed a deep learning-based model for classifying expression of ER, PR and HER2. Panambur et al.[37] used a ResNet-18-based model with transfer learning to classify the luminal versus non-luminal subtypes in full mammogram images. Zhou et al.[38] used The LASSO method to select the optimal predictive features and support vector machine and logistic regression models to evaluate the HER-2 status in breast cancer patients. Son et al.[39] predicted four molecular subtypes of breast cancer using radiomics signatures extracted from synthetic mammography reconstructed from digital breast tomosynthesis (DBT). They used features extracted by the first-order statistics, gray level co-occurrence matrix (GLCM), gray level run length matrix (GLRLM), gray level size zone matrix (GLSZM) and the elastic-net approach as a feature selection method. Table 8 illustrates a summary of some related works, including their working methods, used datasets and results.

Despite the numerous advantages of deep learning methods in the field of automated feature extraction and high-level feature learning, few researchers have worked on deep learning-based models for breast cancer molecular subtypes classification using mammography images. Therefore, we propose a novel hierarchical classification strategy, named, the two-stage classification strategy based on deep learning for breast cancer molecular subtypes classification. Additionally, to develop safe and reliable models for breast cancer molecular subtypes classification, we quantify uncertainty in the predictions using the Monte Carlo Dropout technique as approximate Bayesian inference.

### 3. Materials and method

#### 3.1. Datasets

In this study, a public dataset containing 3712 full mammogram images from 1775 patients, named the Chinese Mammography Database (CMMD), was used which is divided into two branches. The second part of the dataset includes 1498 mammographies for 749 patients with known molecular subtypes, including luminal A, luminal B, HER2-enriched, and Triple-negative. For each subject, craniocaudal (CC) and mediolateral oblique (MLO) projection images were obtained. In addition to images and labels of Immunohistochemistry results, the dataset contains other clinical information including the age of the patient, the type of abnormalities including calcification and mass and biopsy confirmed label of benign or malignant tumors[40].

The number of patients for each breast cancer molecular subtype is shown in Fig.1 (a) and each subtype with respect to age is shown in Fig.1 (b). The ages were frequently between 21 and 87.

All 749 patients who had known molecular subtypes had mammography images available in two directions, mediolateral oblique (MLO) and craniocaudal (CC). Previous works have shown that the combination of these two views achieved higher performance[12, 38]. Therefore, for increasing the number of images and achieving better performance, both MLO and CC images were used. Fig.2 shows some random image samples from the used dataset.

### 3.2. Preprocessing

In this section, pre-processing steps are described in order.

1. **Splitting:** First, the dataset images are randomly split into training and testing groups by 80% and 20%, respectively.
2. **Resizing and Normalization:** In this step, each image was resized to $224 \times 224 \times 3$ and its intensity values were normalized to the interval [0, 1].
3. **Solving the data imbalance problem:** In order to overcome the imbalanced number of lesions belonging to each category, specifically in HER2-enriched and triple negative, the Adaptive Synthetic (ADASYN) and random over-sampling algorithms[41, 42] were applied to augment the dataset and improve the classification performance. The various oversampling techniques have been used in many previous works to address a similar data imbalance problem for image classification [12, 39, 43, 44]. Because of better performance, the ADASYN algorithm was applied to the triple negative subtype images and the random over-sampling algorithm was applied the HER2-enriched subtype images.
4. **Data augmentation:** Image augmentation is the process of increasing the number of image training samples, by applying some transformations to them in order to address the over-fitting problem caused by the small training datasets[45].
   Previously, over-sampling techniques were used to increase the number of images in training set. After increasing the number of image samples, the augmentation methods including image rotation within the range of $(-90°, 90°)$ and flipping horizontally and vertically were applied using the API Keras Augmentor.

### 3.3. Two-stage classification strategy

Breast cancer molecular subtypes classification is a multiclass classification task. First, we applied various deep learning models to classify the molecular subtypes with three classes, as a conventional multi-class classification task. The obtained results were not proper, as shown in Table 1. Some studies showed the binary classification of breast cancer molecular subtypes using mammograms has relatively proper results in which they often used machine or deep learning algorithms. This binary classification includes triple-negative vs non–triple-negative, HER2-enriched vs non–HER2-enriched, and luminal vs non-luminal subtypes. However, they do not correspond to the multi-label essence of the breast cancer molecular subtypes and they can only separate the classes one by one and they cannot assign one of the three labels to an image.

In this study, we propose a two-stage classification strategy to improve the performance of the breast cancer molecular subtypes classification task. The proposed two-stage classification strategy composes two binary classification tasks. The first task classifies triple negative vs non–triple negative cases, while the second binary classification task classifies luminal vs non-luminal subtype. The cases of HER2-enriched are classified as the non-luminal output of the second binary classifier. The two-stage classification strategy includes two independent deep learning models and training lines. For better understanding, a summary of the proposed two-stage classification strategy is shown in Fig.3.

### 3.4. Transfer learning and fine-tuning

Transfer learning is a method that uses a pre-trained model on a large amount of data, and then transfers the pre-trained model to the smaller studied datasets for fine-tuning. In this study, we performed transfer learning methods based on some popular pre-trained CNN models that were trained with the ImageNet dataset and retrained the models using the considered dataset for fine-tuning.

The proposed two-stage classification strategy, as mentioned before, includes two independent deep learning models and training lines. Therefore, various pre-trained deep learning models were applied to choose the best models for the two binary tasks.

### 3.5. Learning methods

In the two-stage classification strategy, we have two independent training lines for two binary classification tasks. So, some of the hyperparameters were different for each training line. The ADAM optimizer is used for both binary tasks, with an initial learning rate of 0.0001 and batch size of 32. Also, L2 regularization with the regularization parameter 0.000001 is used for the classifier layers. The binary cross-entropy loss function was used to select the optimal weights. Additionally, the number of epochs is 55 for training the first Model and 75 for training the second Model.

### 3.6. Uncertainty Quantification method

In this study, the Monte Carlo (MC) Dropout method is used to quantify predictive uncertainty. The MC Dropout method is a principled and practical way for quantifying both types of uncertainties, namely, aleatoric and epistemic uncertainty, without sacrificing either computational complexity or model performance.

Dropout is a regularization technique that is widely used to solve over-fitting problems in deep neural networks by randomly disabling some neurons with a dropout probability, p [46]. During the training process, dropout is applied, hence, a different subset of network architecture is made to prevent excessive co-tuning using stochastic forward passes[27].

In the Monte Carlo Dropout method, as a Bayesian approximation method, the dropouts apply not only during training but also at test time. So, we can obtain the predictive distribution by applying the same input with collecting T stochastic forward passes and averaging them, as shown in eq. (1):

$$p(y^* = c|x^*, X, Y) \approx \frac{1}{T} \sum_t p(y^* = c|x^*, w_t) \qquad (1)$$

Where c is all possible classes that y can take, X, Y training inputs and their corresponding outputs, respectively, $x^*, y^*$ a pair of input and output test sample, $w_t \sim q(w)$, $q(w)$ denotes a dropout variational distribution and $p(y^* = c|x^*, w)$ is the SoftMax function which is written in eq. (2):

$$p(y^* = c|x^*, w) = \frac{\exp(f_c^w)}{\sum_{c'} \exp(f_{c'}^w)}$$

$$\quad (2)$$

Where $f_c^w$ denotes a network function parametrized by the variables w[31, 47]. Therefore, by using the predictive distribution, we can quantify the predictive uncertainty through the predictive entropy [46] in the deep learning models, as defined as eq. (3):

$$H_{p(y^*|x^*,X,Y)}[y^*] = -\sum_{y^*=c} p(y^* = c|x^*,X,Y) \log p(y^* = c|x^*,X,Y) \quad (3)$$

## 4. Results

### 4.1. Experimental setting

To implement the models, we used Python 3, TensorFlow and Keras framework. All the experiments have been executed on the NVIDIA Tesla T4 GPU with 16 GB memory configured machine.

### 4.2. Performance metrics

The performance of the proposed breast cancer molecular subtypes classification task was evaluated based on several metrics: accuracy, recall (Sensitivity), precision and F1 score. Also, the area under the curve (AUC) was used to evaluate each class individually. The calculation formulas are written in eq. (4-7). Correspondingly, the confusion matrix and the ROC curves were introduced for the proposed model.

$$\text{Accuracy} = \frac{TP + TN}{TP + TN + FP + FN} \quad (4)$$

$$\text{Recall} = \frac{TP}{TP + FN} \quad (5)$$

$$\text{Precision} = \frac{TP}{TP + FP} \quad (6)$$

$$\text{F1 score} = 2 \times \frac{\text{Recall} \times \text{Precision}}{\text{Recall} + \text{Precision}} = \frac{2 \times \text{TP}}{2 \times \text{TP} + \text{FP} + \text{FN}} \tag{7}$$

Where TP, TN, FN and FP denote True Positive, True Negative, False Negative, and False Positive rates.

### 4.3. Uncertainty quantification metrics

As mentioned before, we quantified the uncertainty of the classification predictions by predictive entropy defined as eq. (3). Due to the predictive entropy capturing epistemic and aleatoric uncertainty, this quantity is high when either the aleatoric or epistemic uncertainty is high.

### 4.4. Breast cancer molecular subtypes classification results

First, we applied various deep learning models to classify three molecular subtypes, as a conventional multi-class classification task. The obtained results are shown in Table 1. The best-performing network has been reported for EfficientNetB4 model with an averaged precision, recall, F1-score and accuracy of 58.49%, 56.42%, 52.71%, and 54.57%, respectively. However, these obtained results were not satisfactory. Therefore, in this study, we propose a two-stage classification strategy to improve the performance of the breast cancer molecular subtypes classification task.

As mentioned earlier, two different experiments were implemented for the classification of HER2-enriched, luminal and triple-negative subtypes using the two-stage classification strategy. In this section, we present the results for each proposed experiment, followed by a brief analysis.

With the aim of determining the best models for two binary classification tasks, additional experiments have been carried out using some popular Deep learning models. As shown in Table 2, the InceptionResNetV2 achieved the best results with an averaged precision, recall, F1-score and accuracy of 87.90%, 86.42%, 86.29%, and 86.42%, respectively for the first model. For the second model, the EfficientNetB4 obtained the best results with an averaged precision, recall, F1-score and accuracy of 65.00%, 69.84%, 63.98%, and 67.10%, respectively.

Therefore, we combined two selected models to construct the hierarchical two-stage classification strategy. The first model includes the InceptionResNetV2 with a classifier block to classify the image into triple negative vs non–triple negative classes. The outputs of the first model are entered into the second model in which the EfficientNetB4 and a classifier block classify luminal vs non-luminal subtypes. Therefore, in the first step, the triple negative cases are extracted and in the second step, the luminal and the HER2-enriched cases are predicted.

The classifier blocks include a stack of three Fully-Connected and two Monte Carlo Dropout layers: the first two Fully-Connected layers have 4096 neurons each with a ReLu activation function, and the third one performs binary classification and thus, includes two neurons with a SoftMax activation function . To quantify the uncertainty, there is also a Monte Carlo Dropout layer between each Fully-Connected layer in the proposed models.

As shown in Table 3, the proposed model provides an average precision, recall, F1-score and accuracy of 67.43%, 67.65%, 66.52%, and 69.88%, respectively. The AUC is 0.70, 0.74 and 0.86 for HER2-enriched, luminal and triple-negative subtypes. Also, Fig.4(a) and Fig.4 (b) show ROC curves and the confusion matrix for the proposed model. The results of the proposed model show a significant increase of performance compared to the conventional multi-class classification methods.

One of the main objectives of this study is the quantification of the uncertainty with the aim of increasing the reliability of the proposed model. So, we used the Monte Carlo Dropout method as a Bayesian approximation to quantify the uncertainty. With considering uncertainty, Table 3 illustrates an averaged precision, recall, F1-score and accuracy of 67.88%, 68.11%, 66.94%, and 70.25%, respectively. Also, the AUC is 0.71, 0.75 and 0.86 for HER2-enriched, luminal and triple-negative subtypes. The results show that the model with uncertainty quantification method, as a Bayesian deep learning, has had slightly better performance than the proposed model without uncertainty quantification. Also, Fig. 5(a) and Fig.5 (b) show that the ROC curves and the confusion matrix of the uncertainty-aware model for the different classes of molecular subtypes. In general, the results show promising classification performance of the Bayesian deep learning and indicate that the proposed model did not sacrifice performance while can estimate the uncertainty values. Also, Fig.6-8 show some images with low and high uncertainty values.

5. **Discussion**

In the present work, a reliable novel hierarchical classification strategy, named, the two-stage classification strategy was proposed for predicting the three-class classification (HER2-enriched, luminal and triple-negative) of breast cancer molecular subtypes using the full mammogram images. The proposed model was compared with the conventional multi-class classification methods, according to averaged accuracy, recall, precision, F1 score, as well as per-class AUC. The results showed that the proposed model had better performance than the conventional multi-class classification methods. Also, we demonstrated that the Monte Carlo method in the proposed model, not only quantifies the uncertainty, but also increases the performance. Consequently, the uncertainty-aware proposed model achieved an average accuracy of 70.25%, recall of 68.11%, precision of 67.88%, and F1-score of 66.94%. The per-class AUC was 0.71, 0.75 and 0.86 for the HER2-enriched, luminal and triple-negative classes, respectively .

Additionally, the proposed model was compared to some existing studies. As shown in Table 4, the proposed strategy is not only comparable to some works that used various modalities, but also due to quantify the uncertainty using the Mc dropout method, the proposed model is more reliable than others, even though it used a limited number of full mammogram images.

On the other hand, the proposed model has proper performance in the detection of the triple negative subtype, which has stronger invasiveness, higher recurrence and metastasis rate, and lower survival rate for patients. Therefore, it can be said that our model has the prognostic value.

One of the limitations of this study is that the used dataset is imbalanced. The number of images in HER2-enriched and triple-negative subtypes is much lower than in luminal cases. Therefore, to address the imbalanced issue, in this study, the oversampling methods were used. Also, the ROI of the images is not marked. Although the results are promising, they may improve with more balanced images and corresponding ROIs.

## 6. Conclusion and future work

Breast cancer is the most commonly diagnosed cancer in women. It is a heterogeneous disease with different molecular subtypes, clinical behavior, treatment responses as well as survival outcomes. The immunohistochemistry methods that are mainly used to detect receptor expression status, rely on small tissue samples and do not represent the complete features of the tumor. There is a temporal heterogeneity of breast tumors. Additionally, it is an expensive test, and not readily available in many developing and under-developed countries. Therefore, there is a need for the development of a reliable, accurate, available and inexpensive alternative method to predict the molecular subtypes of breast cancer using medical images and provide personalized cancer treatment plans.

In this study, a novel hierarchical classification strategy, named, the two-stage classification strategy was proposed for the prediction of three-class classification (HER2-enriched, luminal and triple-negative)breast cancer molecular subtypes using the full mammogram images. The proposed model is a combination of the InceptionResNetV2 and the EfficientNetB4 models with different kinds of training lines.

On the other hand, a Computer-Aided Diagnosis or Prognosis system must be reliable. The predicted results from a Computer-Aided Diagnosis or Prognosis system could introduce biases affecting the judgment of the expert. Hence, the predictive uncertainty quantification is an immediate priority to make a reliable deep learning based automated disease diagnosis/prognosis system. Therefore, we used the Monte Carlo dropout method as a Bayesian approximation to quantify the predictive uncertainty values. In addition to the advantage of uncertainty quantification, this method also increased the performance. Consequently, the uncertainty-aware proposed model achieved per-class AUCs of 0.71, 0.75 and 0.86 for HER2-enriched, luminal and triple-negative classes, respectively. Furthermore, the proposed model has a proper performance in detecting of the triple negative subtype, which has stronger invasiveness, higher recurrence and metastasis rate, and lower survival rate of patient. Thus, it can be concluded that our model has the prognostic value.

Future work will attempt to enhance the performance of the proposed model by using attention-based techniques.

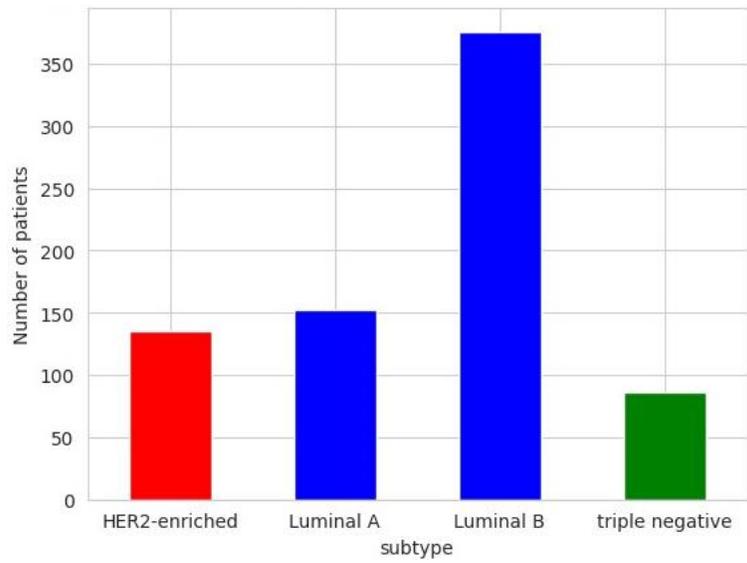

(a)

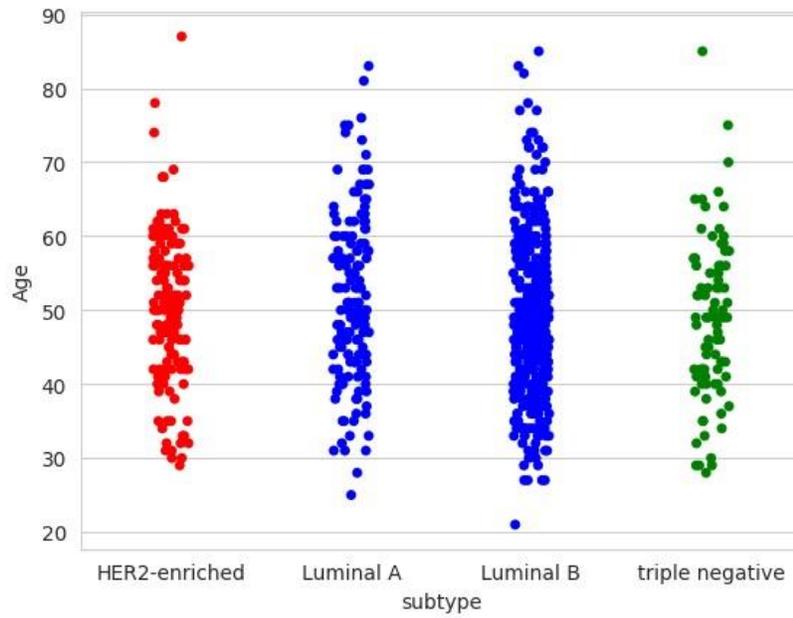

(b)

Fig.1. (a) The number of patients for each breast cancer molecular subtypes and (b) each subtypes respect to the ages.

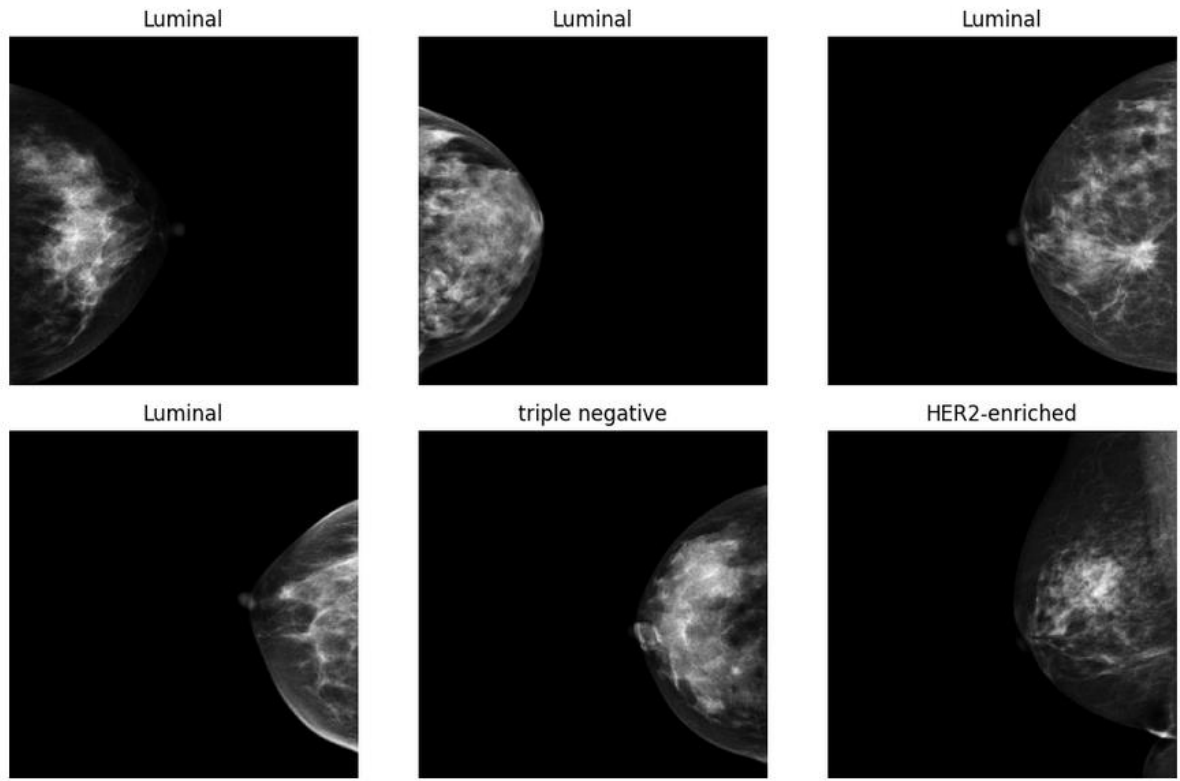

Fig.2. Some random Mammography samples from the used datasets with corresponding molecular subtype labels.

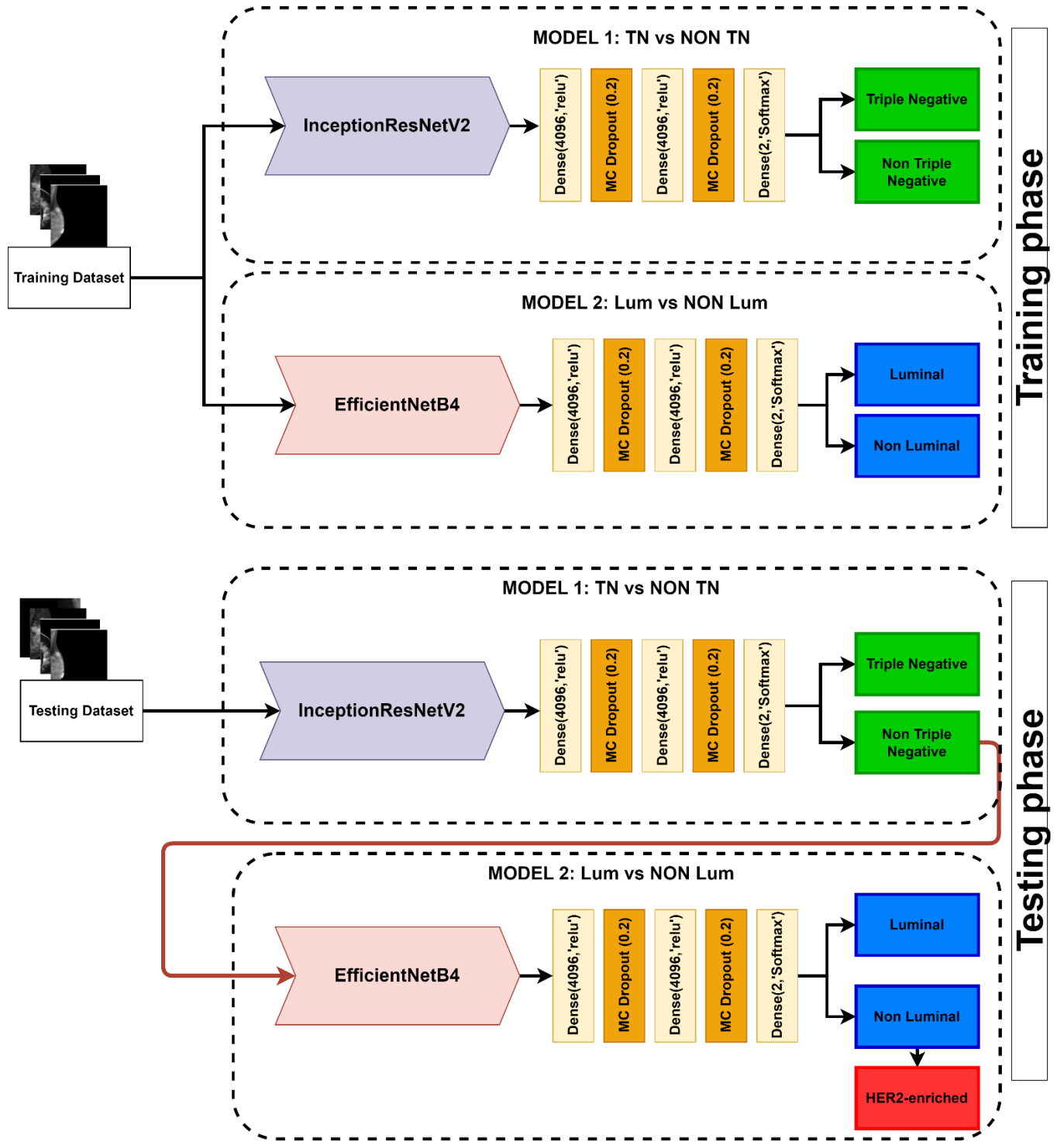

Fig.3. Summary of the proposed two-stage classification strategy.

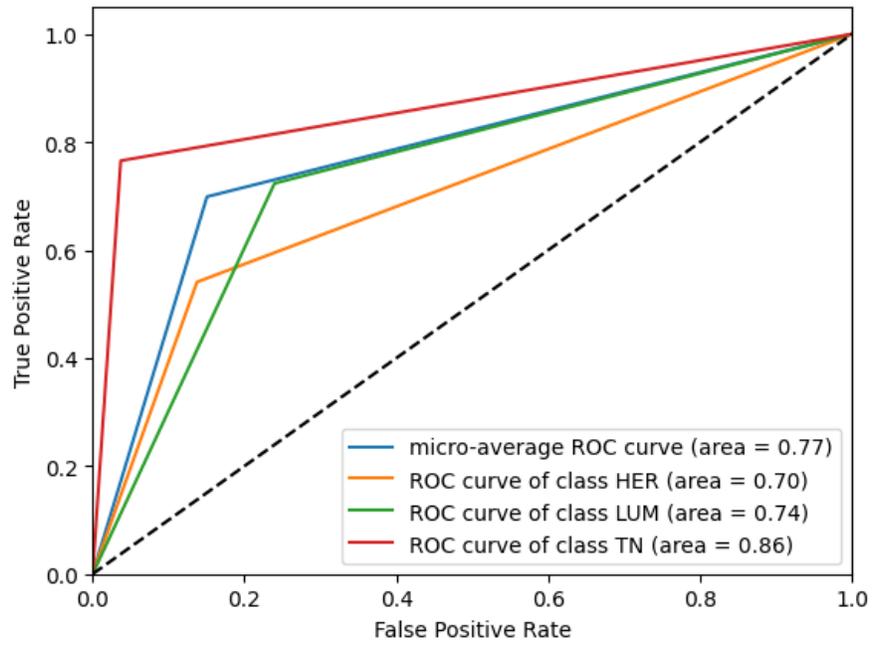

(a)

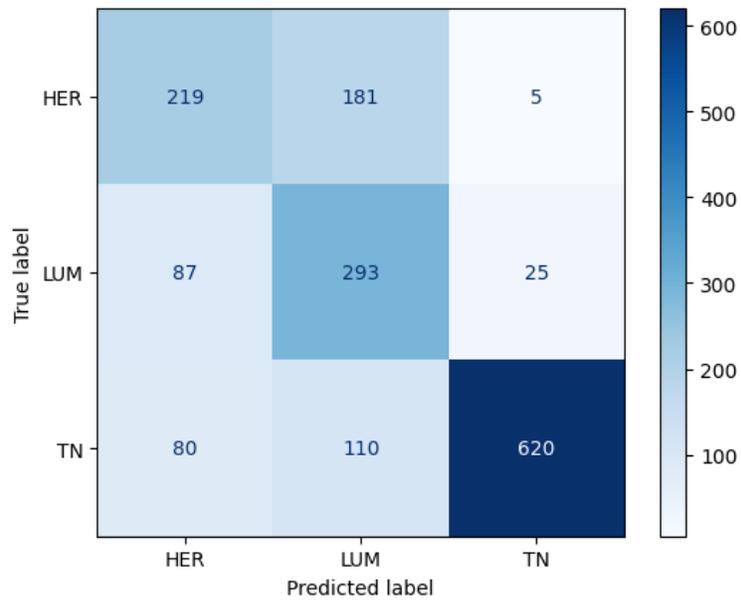

(b)

Fig.4. (a) ROC curves and (b) confusion matrix results obtained for the proposed two-stage classification model without considering uncertainty (non Bayesian model)

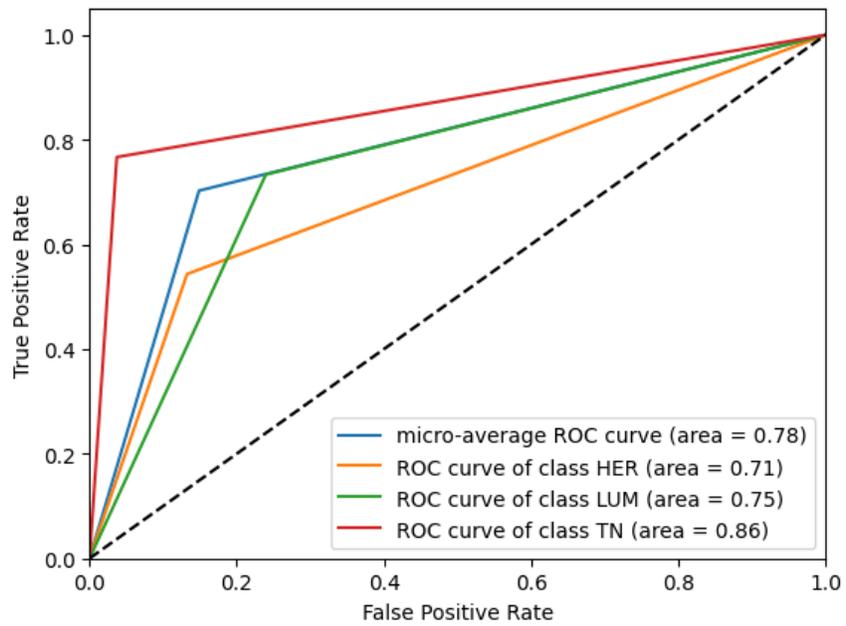

(a)

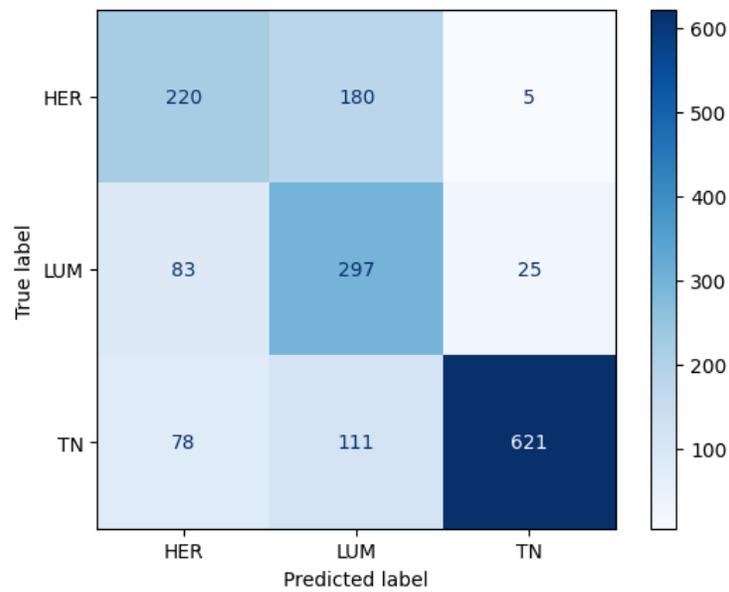

(b)

Fig.5. (a) ROC curves and (b) confusion matrix results obtained for the proposed two-stage classification model with considering uncertainty (Bayesian model)

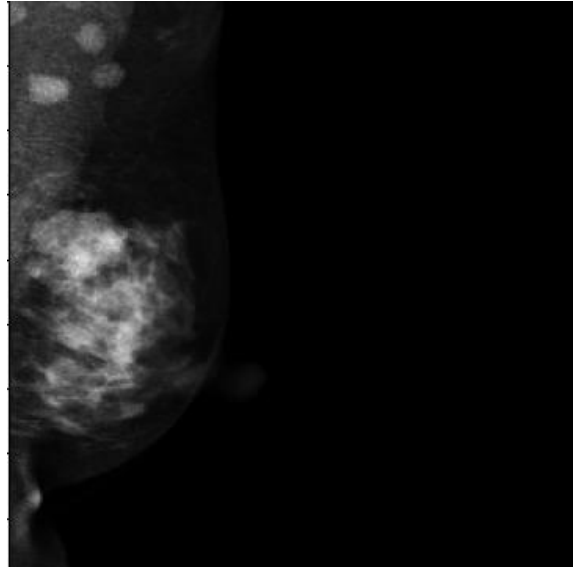

(a)

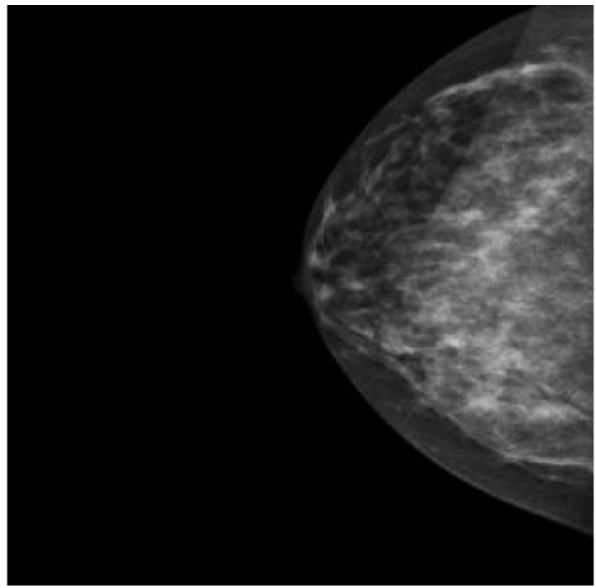

(b)

Fig. 6. (a) An image with low uncertainty ( predictive entropy=0.0012) and (b) an image with high uncertainty ( predictive entropy=0.6930) in the triple-negative class.

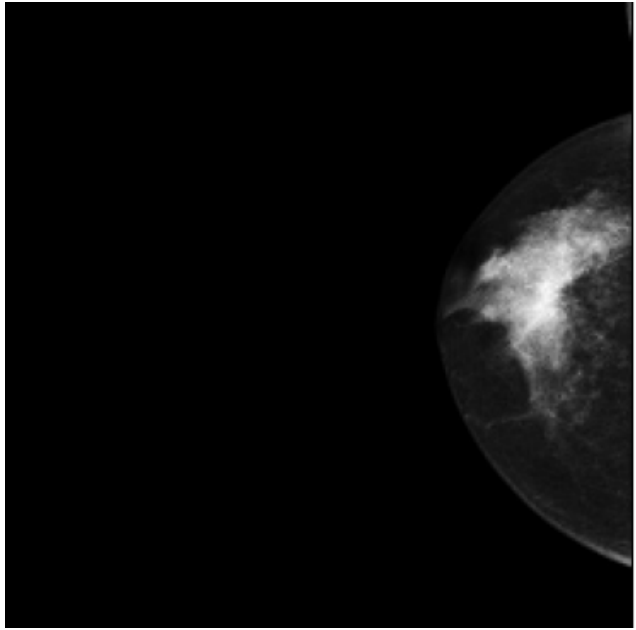

(a)

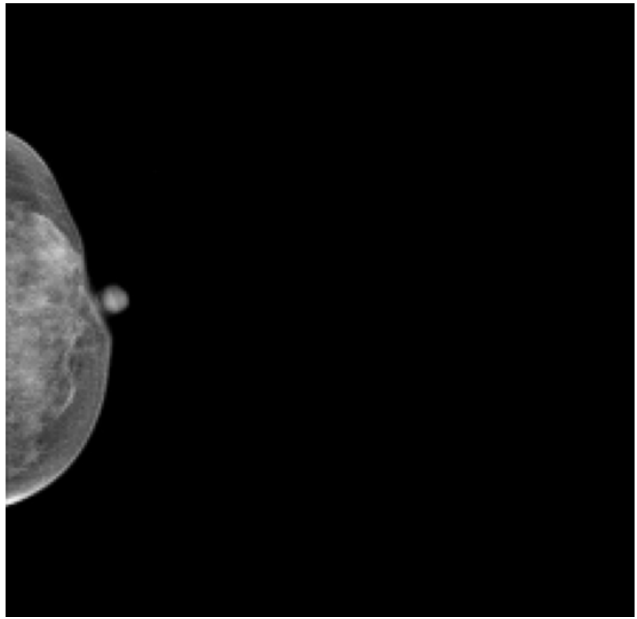

(b)

Fig. 7. (a) An image with low uncertainty ( predictive entropy=0.0018) and (b) an image with high uncertainty ( predictive entropy=0.6845) in the Luminal class.

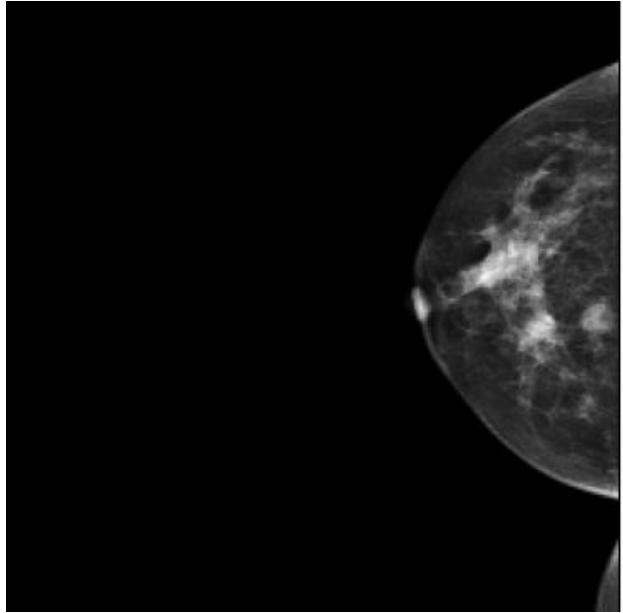

(a)

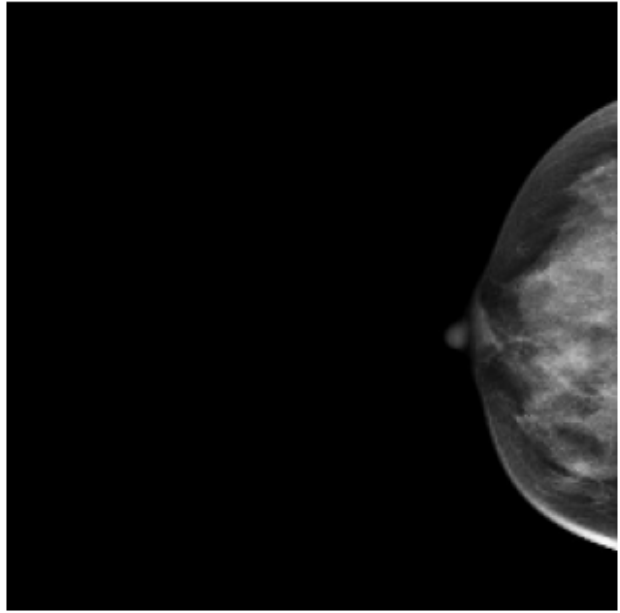

(b)

Fig. 8. (a) An image with low uncertainty ( predictive entropy=0.0019) and (b) an image with high uncertainty ( predictive entropy=0.6699) in the HER2-enriched class.

**Table 1.** Conventional Multi-class classification performance of various deep learning models (epoch=30).

| Model | Precision (%) | Recall (%) | F-Measure (%) | Accuracy (%) | AUC | | |
|---|---|---|---|---|---|---|---|
| | | | | | HER2-enriched | luminal | triple-negative |
| VGG16 | 16.67 | 33.33 | 22.22 | 50.00 | - | - | - |
| DenseNet121 | 44.02 | 46.58 | 43.79 | 51.36 | 0.50 | 0.66 | 0.65 |
| InceptionV3 | 50.39 | 50.29 | 44.81 | 48.40 | 0.51 | 0.71 | 0.67 |
| InceptionResNetV2 | 49.39 | 49.67 | 45.84 | 48.15 | 0.52 | 0.70 | 0.66 |
| **EfficientNetB4** | **58.49** | **56.42** | **52.71** | **54.57** | **0.60** | **0.71** | **0.73** |

**Table 2.** The results of the first (Model1: triple-negative vs non-triple-negative) and the second (Model2: luminal vs non-luminal ) Binary classification tasks of various deep learning models.

|  |  |  | Precision (%) | Recall (%) | F-Measure (%) | Accuracy (%) |
|---|---|---|---|---|---|---|
| Model1: triple-negative vs non-triple-negative | | VGG16 | 74.84 | 74.69 | 74.65 | 74.69 |
| | | DenseNet121 | 79.03 | 77.04 | 76.64 | 77.04 |
| | | InceptionV3 | 78.08 | 77.65 | 77.56 | 77.64 |
| | | **InceptionResNetV2** | **87.90** | **86.42** | **86.29** | **86.42** |
| | | EfficientNetB4 | 84.32 | 79.20 | 78.39 | 79.20 |
| Model2: luminal vs non-luminal | | VGG16 | 37.50 | 50.00 | 42.86 | 50.00 |
| | | DenseNet121 | ۵۷/۹۳ | ۵۹/۷۵ | ۴۹/۷۵ | ۵۰/۳۷ |
| | | InceptionV3 | 62.57 | 66.75 | 60.55 | 63.33 |
| | | InceptionResNetV2 | 63.91 | 66.09 | 52.75 | 52.96 |
| | | **EfficientNetB4** | **65.00** | **69.84** | **63.98** | **67.10** |

**Table 3.** The results of the proposed hierarchical model for the prediction of three-class clasification (HER2-enriched, luminal and triple-negative) breast cancer molecular subtypes. (obtained results with and without considering the uncertainty)

|  | Without UQ | | | | | With UQ | | | | |
|---|---|---|---|---|---|---|---|---|---|---|
| Class | AUC | Precision (%) | Recall (%) | F-Measure (%) | Accuracy (%) | AUC | Precision (%) | Recall (%) | F-Measure (%) | Accuracy (%) |
| HER2-enriched | 0.70 | 56.74 | 54.07 | 55.37 | | 0.71 | 57.74 | 54.32 | 55.98 | |
| Luminal | 0.74 | 50.17 | 72.35 | 59.25 | | 0.75 | 50.51 | 73.33 | 59.82 | |
| Triple-negative | 0.86 | 95.38 | 76.54 | 84.93 | | 0.86 | 95.39 | 76.67 | 85.01 | |
| Averaged | 0.77 | 67.43 | 67.65 | 66.52 | 69.88 | 0.78 | 67.88 | 68.11 | 66.94 | 70.25 |

UQ: Uncertainty Quantification

**Table 4.** Comparison of the proposed model with other existing methods for the prediction of multi-class clasification (HER2-enriched, luminal and triple-negative) breast cancer molecular subtypes.

| Study | year | Medical Image | Dataset | Method | AUC | | | UQ |
|---|---|---|---|---|---|---|---|---|
| | | | | | Luminal (A+B) | HER2-enriched | Triple-negative | |
| Ma et al.[12] | 2019 | Mammograms | private | Naive Bayes | 0.75 | 0.78 | 0.87 | No |
| Son et al. [39] | 2020 | Mammograms | private | Elastic-net | 0.65 | 0.56 | 0.84 | No |
| Li et al.[32] | 2022 | Sonograms | private | DenseNet-121 | 0.58 | - | 0.56 | No |
| | | | | EfficientNet-B2 | 0.60 | - | 0.58 | |
| | | | | VGGNet-19 | 0.56 | - | 0.57 | |
| Zhoua et al. [38] | 2019 | Mammograms | private | logistic regression | - | 0.79 | - | No |
| Xu te al. [48] | 2022 | Sonograms (3D) | private | logistic regression | 0.72 | 0.75 | 0.78 | No |
| Ha et al.[49] | 2019 | MRI | private | CNN | 0.83 | 0.88 | 0.92 | No |
| Ours | | Mammograms | CMMD | Two-stage classification stretegy using BDL | 0.75 | 0.71 | 0.86 | Yes |

UQ: Uncertainty Quantification